\documentclass[preprint,3p,twocolumn]{elsarticle}

\usepackage[colorlinks]{hyperref}
\usepackage{amsmath}
\usepackage{bm}
\usepackage{lineno}
\usepackage{mathtools}
\usepackage{subcaption}
\usepackage{pgfplotstable}
\usepackage{siunitx}
\usepackage{graphicx}
\usepackage{booktabs}
\usepackage[utf8]{inputenc}
\usepackage[capitalize,noabbrev]{cleveref}
\usepackage{booktabs}
\usepackage[section]{placeins}
\usepackage{notoccite}

\creflabelformat{equation}{#2#1#3}

\DeclarePairedDelimiter\abs{\lvert}{\rvert}
\DeclarePairedDelimiter\norm{\lVert}{\rVert}

\begin{document}

\begin{frontmatter}

\title{Enhancing Convolutional Neural Networks for Face Recognition with Occlusion Maps and Batch Triplet Loss}

\author[uh,idscan]{Daniel Sáez Trigueros}

\author[uh]{Li Meng\corref{correspondingauthor}}
\ead{l.1.meng@herts.ac.uk}

\author[idscan]{Margaret Hartnett}

\cortext[correspondingauthor]{Corresponding author}

\address[uh]{School of Engineering and Technology, University of Hertfordshire, Hatfield AL10 9AB, UK}
\address[idscan]{IDscan Biometrics (a GBG company), London E14 9QD, UK}

\begin{abstract}
Despite the recent success of convolutional neural networks for computer vision applications, unconstrained face recognition remains a challenge. In this work, we make two contributions to the field. Firstly, we consider the problem of face recognition with partial occlusions and show how current approaches might suffer significant performance degradation when dealing with this kind of face images. We propose a simple method to find out which parts of the human face are more important to achieve a high recognition rate, and use that information during training to force a convolutional neural network to learn discriminative features from all the face regions more equally, including those that typical approaches tend to pay less attention to. We test the accuracy of the proposed method when dealing with real-life occlusions using the AR face database. Secondly, we propose a novel loss function called \textit{batch triplet loss} that improves the performance of the triplet loss by adding an extra term to the loss function to cause minimisation of the standard deviation of both positive and negative scores. We show consistent improvement in the Labeled Faces in the Wild (LFW) benchmark by applying both proposed adjustments to the convolutional neural network training.
\end{abstract}


\end{frontmatter}

\section{Introduction} % (fold)
\label{sec:introduction}
Deep learning models, in particular convolutional neural networks (CNNs), have revolutionised many computer vision applications, including face recognition. As recent benchmarks \cite{Huang2007Labeled,kemelmacher2016megaface,guo2016ms} show, most of the top performing face recognition algorithms are based on CNNs. Even though these models need to be trained with hundreds of thousands of faces to achieve state-of-the-art accuracy, several large-scale face datasets \cite{Yi2014Learning,Parkhi2015Deep,guo2016ms} have recently been made publicly available to facilitate this.

Most of the recent research in the field has focused on unconstrained face recognition. CNN models have shown excellent performance on this task, as they are able to extract features that are robust to variations present in the training data (if enough samples containing these variations are provided). Nonetheless, in this work, we show how partial facial occlusions remain a problem for unconstrained face recognition. This is because most databases used for training do not present enough occluded faces for a CNN to learn how to deal with them. Common sources of occlusion include sunglasses, hats, scarves, hair, or any object between the face and the camera. This is of particular relevance to applications where the subjects are not expected to be co-operative (e.g. in security applications). One way of overcoming this problem is to train CNN models with datasets that contain more occluded faces. However, this task can be challenging because the main source of face images is usually the web, where labelled faces with occlusions are less abundant.

Bearing this in mind, we propose a novel data augmentation approach for generating occluded face images in a strategic manner. We use a technique similar to the occlusion sensitivity experiment proposed in \cite{Zeiler2014Visualizing} to identify the face regions where a CNN extracts the most discriminative features from. In our proposed method, the identified face regions are covered during training to force a CNN to extract discriminative features from the non-occluded face regions with the goal of reducing the model's reliance on the identified face regions. Our CNN models trained using this approach have demonstrated noticeable performance improvement on face images presenting real-life facial occlusions in the AR face database \cite{martinez1998ar}.

CNN models for face recognition can be trained using different approaches. One of them consists of treating the problem as a classification one, wherein each identity in the training set corresponds to a class. After training, the model can be used to recognise faces that are not present in the training set by discarding the classification layer and using the features of the previous layer as the face representation. In the realm of deep learning, these features are commonly referred to as bottleneck features. Following this first training stage, the model can be further trained using other techniques to optimise the bottleneck features for the target application \cite{Yi2014Learning,Sun2014Deep}. Another common approach to learning face representation is to directly learn bottleneck features by optimising a distance metric between pairs of faces \cite{chopra2005learning,Fan2014Learning} or triplets of faces \cite{Schroff2015Facenet}.

Positive results have been demonstrated when combining these two techniques, either by (i) jointly training with a classification loss and a distance metric loss \cite{Sun2014Deepa}; or (ii) by first training with a classification loss and then fine-tuning the CNN model with a distance metric loss \cite{Parkhi2015Deep,Sankaranarayanan2016Tripleta,Sankaranarayanan2016Triplet}. In this work, we adopt the latter approach and use the triplet loss to optimise bottleneck features. The goal of the triplet loss is to separate positive scores (obtained when comparing pairs of faces belonging to the same subject) from negative scores (obtained when comparing pairs of faces belonging to different subjects) by a minimum margin. We argue that training with this loss function can lead to undesired results. Thus, we propose a new loss function to alleviate this issue by also minimising the standard deviation of both positive and negative scores. Using the Labeled Faces in the Wild (LFW) benchmark, we show that the CNN models trained with our proposed loss function consistently outperform those trained with the triplet loss function.

The remainder of this paper is organised as follows. \Cref{sec:related_work} provides a review of the related work, with a focus on deep learning approaches and face recognition with occlusion. \Cref{sec:proposed_methods} details our CNN architecture, training procedure and (i) our method of improving recognition of partially occluded faces; and (ii) our novel loss function. \Cref{sec:experiments} describes our experimental results, and our conclusions are presented in \cref{sec:conclusions}.

% section introduction (end)

\section{Related Work} % (fold)
\label{sec:related_work}

One of the first successful applications of convolutional neural networks was handwritten character recognition \cite{lecun1998gradient}. Soon after, the first face recognition algorithm that included a CNN was proposed in \cite{Lawrence1997Face}. However, unlike \cite{lecun1998gradient}, their algorithm \cite{Lawrence1997Face} was not entirely based on neural networks. Years later, \cite{chopra2005learning} proposed an end-to-end Siamese architecture trained with a contrastive loss function to directly minimise the distance between pairs of faces from the same subject while increasing the distance between pairs of faces from different subjects. These CNN-based face recognition models did not achieve groundbreaking results, mainly due to the low capacity of the networks used and the relatively small datasets available for training at the time. It was not until these models were scaled up and trained with large amount of data \cite{Krizhevsky2012Imagenet} that CNNs became the state-of-the-art approach for face recognition.

In particular, Facebook's DeepFace \cite{Taigman2014Deepface}, one of the first CNN-based approaches for face recognition that used a high capacity model, achieved an accuracy of 97.35\% on the LFW benchmark, reducing the error of the previous state-of-the-art by 27\%. DeepFace used an effective 3D alignment algorithm to frontalise faces before feeding them to a CNN with several convolutional, max-pooling and locally connected layers. The CNN was trained with a dataset containing 4.4 million faces from 4,030 subjects. Concurrently, the DeepID system \cite{Sun2014Deep} achieved similar results by concatenating the bottleneck features of 60 CNNs trained on different face crops and optimising the concatenated feature vector using the Joint Bayesian method proposed in \cite{Chen2012Bayesian}. More work by the same authors \cite{Sun2014Deepa} achieved further performance improvements by simultaneously training with a contrastive loss (similar to the one used in \cite{chopra2005learning}) and a classification loss. The authors claimed that the contrastive loss reduced intra-personal variations and the classification loss increased inter-personal variations. The described system achieved an accuracy of 99.15\% on the LFW benchmark using a relatively small training set containing 202,599 face images of 10,177 identities.

As shown in \cite{Zhou2015NaiveDeep}, training data is one of the most important factors for increasing the accuracy of CNN-based approaches. In particular, it was shown that a CNN model becomes more accurate as the number of different identities in the training set increases, provided that several samples per identity are available. A good example is Google's FaceNet \cite{Schroff2015Facenet}, which used between 100 million and 200 million face images of about 8 million different people for training. A triplet loss function with a novel online triplet sampling strategy was used for training FaceNet, which achieved an accuracy of 99.63\% on the LFW benchmark. The triplet loss has been subsequently used to fine-tune CNNs pre-trained with a classification loss with good results \cite{Parkhi2015Deep,Liu2015Targeting}. Indeed, the triplet loss has become one of the most popular training objectives for face verification \cite{Schroff2015Facenet,Parkhi2015Deep,Liu2015Targeting,Sankaranarayanan2016Tripleta,Sankaranarayanan2016Triplet}, and has been used in other image similarity tasks such as ranking images \cite{Wang2014Learning,hoffer2015deep,Gordo2016Deep} and learning local image descriptors \cite{wohlhart2015learning,kumar2016learning}. Other popular tricks to improve the performance of CNN-based face recognition include Joint Bayesian \cite{Yi2014Learning,chopra2005learning,Chen2012Bayesian,Ding2015Robust} and building ensemble models trained on different face crops \cite{chopra2005learning,Chen2012Bayesian,Liu2015Targeting,Ding2015Robust}.

Recognition of faces with occlusions has been typically handled using two different types of methods, namely, (i) methods that extract local features from the non-occluded regions or (ii) methods that attempt to reconstruct occluded regions.

In the first type of methods, occluded regions are detected first and discarded from the set of local regions used to represent a face. For example, Gabor wavelet features, PCA and SVM were used in \mbox{\cite{Min2011Improving}} to detect the occluded regions and LBP descriptors were used to match the non-occluded regions. In \mbox{\cite{Sharma2013Efficient}}, eigen decomposition was used to generate a reformed image which was subtracted from the original occluded image to locate the occluded regions. Gabor wavelet features and PCA were used to extract features from the non-occluded regions. The method in \mbox{\cite{Park2015Partially}} proposed to extract histograms of Gabor-LBP features on the entire image and then use SIFT keypoint matching to select which subregions should be taken into consideration.

Among the methods that attempt to reconstruct occluded regions, the sparse representation-based classification (SRC) proposed in \mbox{\cite{Wright2009Robust}} has received a lot of attention. This method attempts to represent an occluded test image by a linear combination of training images of the same class and an error term that accounts for the occluded region. The class that gives the closest reconstruction of the original image is considered the correct one. Several improvements to this method have been proposed. For example, \mbox{\cite{Zhou2009Face}} extended SRC by using a Markov random field to model the prior assumption about the spatial continuity of the occluded regions. In \mbox{\cite{jia2008face}} it was proposed to weight each pixel in the image independently to achieve better reconstructed images. Another improvement \mbox{\cite{Yang2013Gabor}} proposed to use linear combinations of Gabor wavelet features instead of pixel intensities, which increased the discrimination power of the face representation and reduced computational costs. The drawback of these methods is that the reconstruction can only be achieved for images of the same class as the training images.

Another method that has gained popularity in image reconstruction tasks such as image denoising and image inpainting is the denoising autoencoder \mbox{\cite{vincent2010stacked,xie2012image}}. The idea is to train a model to learn a mapping between corrupted and clean images. Several approaches have used this idea to reconstruct occluded face images. For example, a stacked sparse denoising autoencoder \mbox{\cite{xie2012image}} with two channels was proposed in \mbox{\cite{Cheng2015Robust}} to discard noise activations in the encoder network and achieve better image reconstructions. Another related method was proposed in \mbox{\cite{zhang2013occlusion}}. They used a novel mapping-autoencoder for occlusion detection and an iterative stacked denoising autoencoder for image reconstruction. More recently, \mbox{\cite{Zhao2016Robust}} proposed to use LSTM autoencoders with two channels to reconstruct faces in the wild. In this method, one autoencoder channel reconstructs the image and the other detects an occlusion mask that is used to replace the occluded region in the original image with the reconstructed pixels. The quality of the final output was further enhanced by introducing an adversarial discriminator.

% section related_work (end)

\section{Proposed Methods} % (fold)
\label{sec:proposed_methods}

We use the CNN architecture proposed in \cite{Yi2014Learning}, which has demonstrated the ability to achieve high accuracy on the LFW benchmark while maintaining low computational complexity. This CNN architecture is similar to that used in \cite{Simonyan2014Very} but comprises only ten convolutional layers and one fully-connected layer. The input to this CNN is a greyscale image of size $100\times100$ pixels aligned using a simple 2D affine transformation. More details about this CNN architecture can be found in \cite{Yi2014Learning}.

As a first training stage, our method adopts the approach of training a classifier wherein the CNN produces a vector of scores $\bm{s}$ for each class $j$, which is passed to a softmax function to calculate the probability $p$ of the correct class $y$:
\begin{equation}
p = \frac{e^{s_y}}{\sum_{j}{e^{s_j}}}\label{eq:softmax_correct_class}
\end{equation}
The total loss of the CNN is defined as the average cross-entropy loss for each training sample $i$:
\begin{equation}
L = -\sum_i^N{\log{p_i}}\label{eq:cross_entropy_class}
\end{equation}
where $N$ is the number of samples in a batch of training samples.

In order to use the trained CNN classification model to compare face images that are not present in the training set, the classification layer (i.e. the layer producing the scores $\bm{s}$) is discarded and the features from the previous layer are used as bottleneck features. These bottleneck features can directly be used as the feature vector representing a face or can be further optimised as described in \cref{sub:batch_triplet_loss}. We have adopted cosine similarity to compare pairs of feature vectors to get a similarity score that indicates the likelihood of two face images belonging to the same identity.

We trained such a CNN classification model using the CASIA-WebFace database \cite{Yi2014Learning}. This database contains 494,414 face images of 10,575 different celebrities gathered from the Internet. We randomly selected 10\% of the images as validation images and used the rest as training images. We consider this CNN model as the baseline for performance comparison in our work and refer to it henceforth as model A.

\subsection{Occlusions Maps} % (fold)
\label{sub:occlusion_maps}

As shown in \cite{Zeiler2014Visualizing}, it is possible to use visualisation techniques to gain insight into the behaviour of CNN models after they have been trained. To solve the facial occlusion challenge, we are interested in identifying which face regions a CNN model relies on the most, as we want to avoid this reliance. Using a classification model, one way of visualising these regions is by observing how a correct class score fluctuates when different face regions are occluded. A similar type of occlusion sensitivity experiment has been conducted in \cite{Zeiler2014Visualizing} in the context of object recognition. In our case, by occluding a face image for which a CNN model predicts the correct class, we can generate a \textit{binary occlusion map} $\bm{O^I}$ to indicate whether placing an occluder at a particular spatial location in the image $\bm{I}$ would cause the model to predict an incorrect class. More formally, a binary occlusion map $\bm{O^I}$ is defined as follows:
\begin{equation}
O^{\bm{I}}_{i,j}=
\begin{cases}
  0, & \text{if}\ \hat{y}_{i,j} = y \\
  1, & \text{otherwise}
\end{cases}
\label{eq:binary_occlusion_map}
\end{equation}
where $\hat{y}_{i,j}$ is the predicted class when the centre of an occluder is placed at the location $(i, j)$ of the image $\bm{I}$ and $y$ is the correct class for the image $\bm{I}$.

\begin{figure}[tb]
\centering
\hspace*{\fill}
\subcaptionbox{\label{fig:map_a}}{\includegraphics[width=0.2\linewidth]{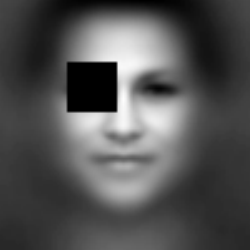}}\hfill
\subcaptionbox{\label{fig:map_b}}{\includegraphics[width=0.2\linewidth]{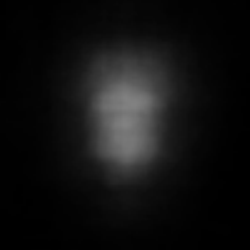}}\hfill
\subcaptionbox{\label{fig:map_c}}{\includegraphics[width=0.2\linewidth]{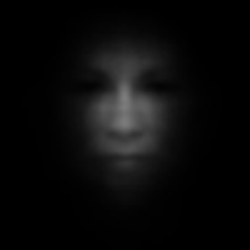}}
\hspace*{\fill}
\par\medskip
\hspace*{\fill}
\subcaptionbox{\label{fig:map_d}}{\includegraphics[width=0.2\linewidth]{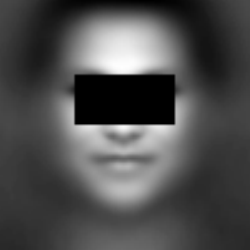}}\hfill
\subcaptionbox{\label{fig:map_e}}{\includegraphics[width=0.2\linewidth]{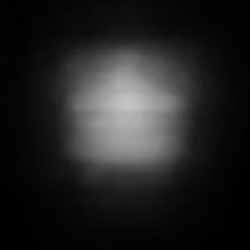}}\hfill
\subcaptionbox{\label{fig:map_f}}{\includegraphics[width=0.2\linewidth]{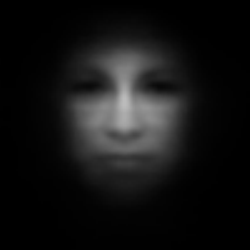}}
\hspace*{\fill}
\par\medskip
\hspace*{\fill}
\subcaptionbox{\label{fig:map_g}}{\includegraphics[width=0.2\linewidth]{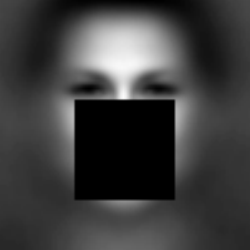}}\hfill
\subcaptionbox{\label{fig:map_h}}{\includegraphics[width=0.2\linewidth]{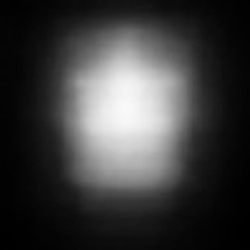}}\hfill
\subcaptionbox{\label{fig:map_i}}{\includegraphics[width=0.2\linewidth]{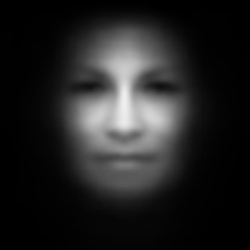}}
\hspace*{\fill}
\caption{\subref*{fig:map_a}, \subref*{fig:map_d}, \subref*{fig:map_g} Mean image occluded at a random location with an occluder of $20\times20$, $20\times40$, and $40\times40$ respectively. \subref*{fig:map_b}, \subref*{fig:map_e}, \subref*{fig:map_h} Occlusion maps $\boldsymbol{O}_{20\times20}$, $\boldsymbol{O}_{20\times40}$, and $\boldsymbol{O}_{40\times40}$ generated using model A and the corresponding occluders. The pixel intensity of the occlusion maps represents the classification error rate when placing the occluder at each location. \subref*{fig:map_c}, \subref*{fig:map_f}, \subref*{fig:map_i} Masked mean image using the occlusion maps $\boldsymbol{O}_{20\times20}$, $\boldsymbol{O}_{20\times40}$, and $\boldsymbol{O}_{40\times40}$ respectively.}
\label{fig:map}
\end{figure}

Since we are using face images that are aligned, we can construct a generic \textit{occlusion map} $\bm{O}$ by simply averaging the binary occlusion maps of a set of face images. Each value of an occlusion map $O_{i,j}$ corresponds to the classification error incurred by a model when an occluder is placed at the location $(i, j)$ in all the images used to generate $\bm{O}$. For convenience, we refer to face regions that present high classification error as \textit{high effect regions} (as these are the regions in which the model relies on the most). By contrast, we refer to face regions that present low classification error as \textit{low effect regions}. These high and low effect regions correspond to the bright and dark areas in the occlusion maps shown in \cref{fig:map} respectively.

Considering the $100\times100$ face images used as input to our model, we experiment with occluders of three different sizes. In particular, we use (i) a square occluder of $20\times20$ pixels that can cover small regions such as one eye, the nose or the mouth as shown in \cref{fig:map_a}; (ii) a rectangular occluder of $20\times40$ pixels that can cover wider regions such as both eyes simultaneously as shown in \cref{fig:map_d}; and (iii) a larger square occluder of $40\times40$ pixels that can cover several face regions simultaneously as shown in \cref{fig:map_g}. We denote the occlusion maps generated with the $20\times20$, $20\times40$ and $40\times40$ occluders by $\bm{O}_{20\times20}$, $\bm{O}_{20\times40}$ and $\bm{O}_{40\times40}$ respectively. \cref{fig:map_b,fig:map_e,fig:map_h} show an example of these occlusion maps generated with model A using 1,000 images from our validation set.

According to \cref{fig:map_c,fig:map_f,fig:map_i}, the central part of the face is one of the highest effect regions. This might be due to the presence of non-frontal face images in the training set. Since the central part of the face is typically visible in both frontal and non-frontal face images, the model learns more discriminative features from this area compared to the outer parts of the face, which might not be visible in non-frontal face images. Simply put, the model is trained with fewer face images in which the outer parts of the face are visible, therefore, it relies more heavily on the central part of the face. We can reverse this behaviour by training with more face images that present occlusions located in high effect regions (central part of the face), as this will force the model to learn more discriminative features from low effect regions (outer parts of the face).

One way of achieving this is by augmenting the training set with face images that present occlusions located at random locations. To do this, during training we can generate occluded training images by overlaying the original training images with a randomly located occluder. However, since we want to favour occlusions in high effect regions, we propose to augment the training set with face images that present occlusions located in high effect regions more frequently than in low effect regions. For this reason, the location of the occluder is sampled from a probability distribution $\bm{P}$ generated by applying the softmax function with the temperature parameter $T$ to an occlusion map $\bm{O}$:
\begin{equation}
P_{i, j} = \frac{e^{\frac{O_{i, j}}{T}}}{\sum_{n, m}{e^{\frac{O_{n, m}}{T}}}}\label{eq:softmax_with_temperature}
\end{equation}
With high temperatures, all locations have the same probability. With low temperatures, locations in high effect regions are assigned a higher probability.

\begin{figure}[tb]
\centering
\hspace*{\fill}
\subcaptionbox{\label{fig:occluder_example_a}}{\includegraphics[width=0.25\linewidth]{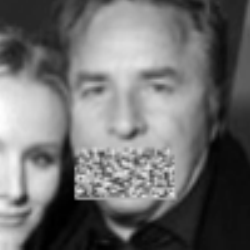}}\hfill
\subcaptionbox{\label{fig:occluder_example_b}}{\includegraphics[width=0.25\linewidth]{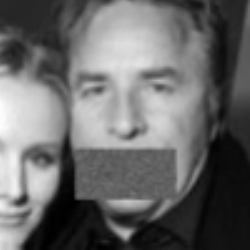}}\hfill
\subcaptionbox{\label{fig:occluder_example_c}}{\includegraphics[width=0.25\linewidth]{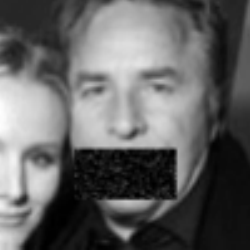}}
\hspace*{\fill}
\caption{Example occluders used during training with different intensities, noise types and noise levels. \subref*{fig:occluder_example_a} Salt-and-pepper noise. \subref*{fig:occluder_example_b} Speckle noise. \subref*{fig:occluder_example_c} Gaussian noise.}
\label{fig:occluder_example}
\end{figure}

As shown in \cref{fig:occluder_example}, we use occluders of random intensities (or random colours if we were dealing with colour images) that present different types of random noise (salt-and-pepper, speckle and Gaussian noise). This is important because if the face is always covered by the same type of occluder, the CNN would only learn features that are robust against that particular type of occlusion. For example, if a black patch is always used to occlude faces during training, the CNN model would perform well when the face is occluded by a black patch, but not when it is occluded by a patch of a different intensity.

This training procedure produces two desired outcomes, namely, (i) the training set is augmented with variations not present in the original data, and (ii) the occluder has a regulariser effect, helping the CNN to learn features from all face regions equally. Both of these increase the generalisation capability of the model and prevent overfitting. In \cref{sec:experiments} we provide experimental results, with both occluded and non-occluded face images, to validate these claims.

% subsection occlusion_maps (end)

\subsection{Batch Triplet Loss} % (fold)
\label{sub:batch_triplet_loss}

In order to make the bottleneck features generalise better to classes not present in the training set, we fine-tune model A using a triplet loss function. This training objective is also used in other similar works \cite{Parkhi2015Deep,Liu2015Targeting,Sankaranarayanan2016Tripleta,Sankaranarayanan2016Triplet}. However, in this work, we fine-tune the bottleneck features directly instead of learning a linear projection from them. It could be argued that the CNN model could be trained from scratch using a triplet loss function, as proposed in \cite{Schroff2015Facenet}. But, according to our experiments, training with softmax cross-entropy loss offers faster convergence than training with a triplet loss when a reasonable number of samples per class are available and the number of classes is not very large.

\begin{figure}[tb]
\centering
\hspace*{\fill}
\subcaptionbox{\label{fig:triplet_a}}{\includegraphics[width=0.35\linewidth]{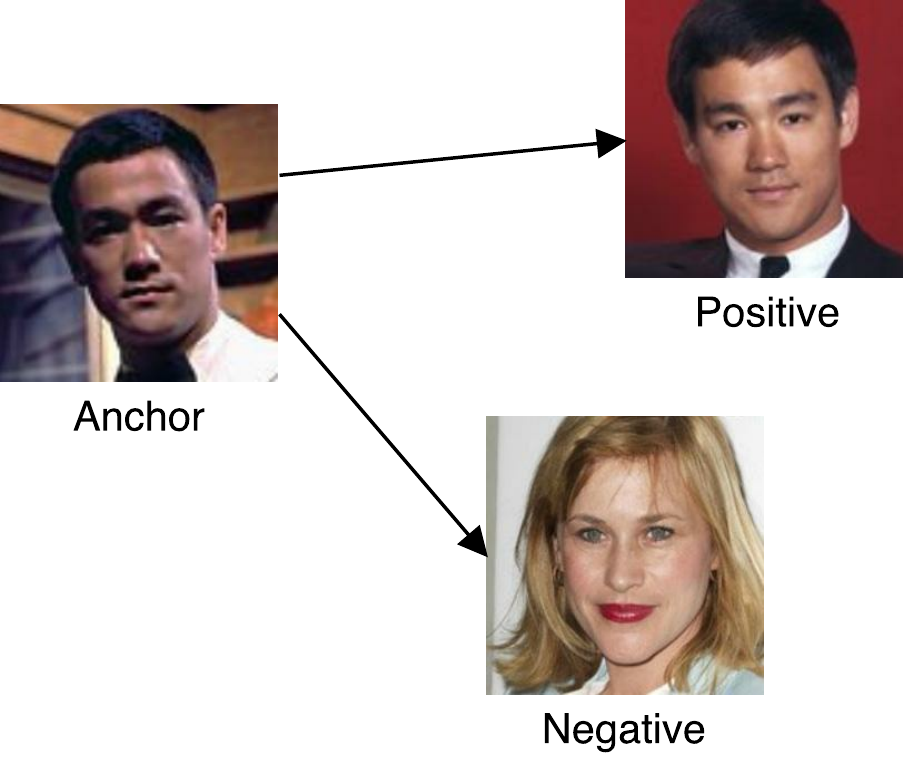}}\hfill
\subcaptionbox{\label{fig:triplet_b}}{\includegraphics[width=0.39\linewidth]{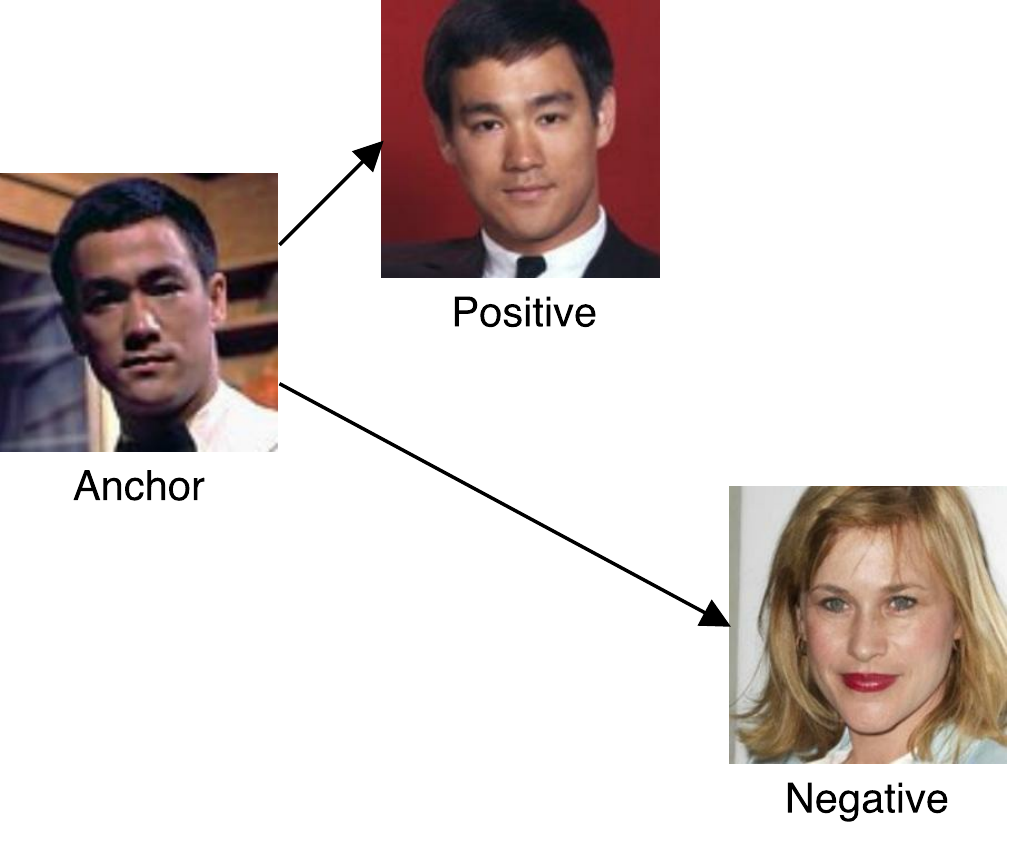}}
\hspace*{\fill}
\caption{Example triplet from the CASIA-WebFace dataset. \subref*{fig:triplet_a} Before triplet training. \subref*{fig:triplet_b} After triplet training.}
\label{fig:triplet}
\end{figure}

To form a triplet we need an anchor image, a positive image and a negative image. The anchor and the positive images belong to the same class and the negative image belongs to a different class. Denoting the output vector of the CNN model as $\bm{z}$ (in our setting this would be the bottleneck features), we can represent the output features for a particular triplet $i$ as $(\bm{z}_i^a, \bm{z}_i^p, \bm{z}_i^n)$, denoting the output features for the anchor, positive and negative images respectively. The goal of a triplet loss function is to make the distance between $\bm{z}_i^a$ and $\bm{z}_i^n$ (i.e. images from different classes) larger than the distance between $\bm{z}_i^a$ and $\bm{z}_i^p$ (i.e. images from the same class) by at least a minimum margin $\alpha$. \cref{fig:triplet} shows a visual representation of a triplet before and after training. In this work, we consider the following as the standard triplet loss function:
\begin{equation}
L = \sum_i^N{\max \left(0, \norm{\bm{z}_i^a-\bm{z}_i^p}_2^2-\norm{\bm{z}_i^a-\bm{z}_i^n}_2^2 + \alpha \right)}\label{eq:standard_triplet_loss}
\end{equation}

Alternative versions of the standard triplet loss function can be defined with distance metrics other than the squared Euclidean distance. For example, the dot product is used as the similarity measure in \cite{Sankaranarayanan2016Tripleta}. More generally, we can write:
\begin{equation}
L = \sum_i^N{\max \left(0, d(\bm{z}_i^a, \bm{z}_i^p)-d(\bm{z}_i^a, \bm{z}_i^n) + \alpha \right)}\label{eq:generic_triplet_loss}
\end{equation}
where $d(\bm{x,y})$ is any function that gives a score indicating distance between two feature vectors. As seen in \cref{eq:generic_triplet_loss}, only triplets that violate the margin condition $d(\bm{z}_i^a, \bm{z}_i^p) + \alpha > d(\bm{z}_i^a, \bm{z}_i^n)$ produce a loss greater than zero and therefore contribute to the model's convergence. To increase the training efficiency, we adopt the online triplet sampling strategy proposed in \cite{Schroff2015Facenet} to select such triplets and only use them during training. Taking this into consideration, we can rewrite \cref{eq:generic_triplet_loss} as:
\begin{equation}
L = \mu_{ap} - \mu_{an} + \alpha\label{eq:simplified_triplet_loss}
\end{equation}
where $\mu_{ap}$ and $\mu_{an}$ are the mean values of the distribution of positive and negative scores respectively.

From \cref{eq:simplified_triplet_loss} we can see that the loss becomes zero whenever $\mu_{an}$ is equal to $\mu_{ap}$ plus the margin $\alpha$. In other words, the triplet loss function tries to separate the mean values of the distribution of positive scores $\mu_{ap}$ from the mean value of the distribution of negative scores $\mu_{an}$ by a minimum margin $\alpha$.

\begin{figure}[tb]
\centering
\subcaptionbox{\label{fig:score_distribution_a}}{\includegraphics[width=0.49\linewidth]{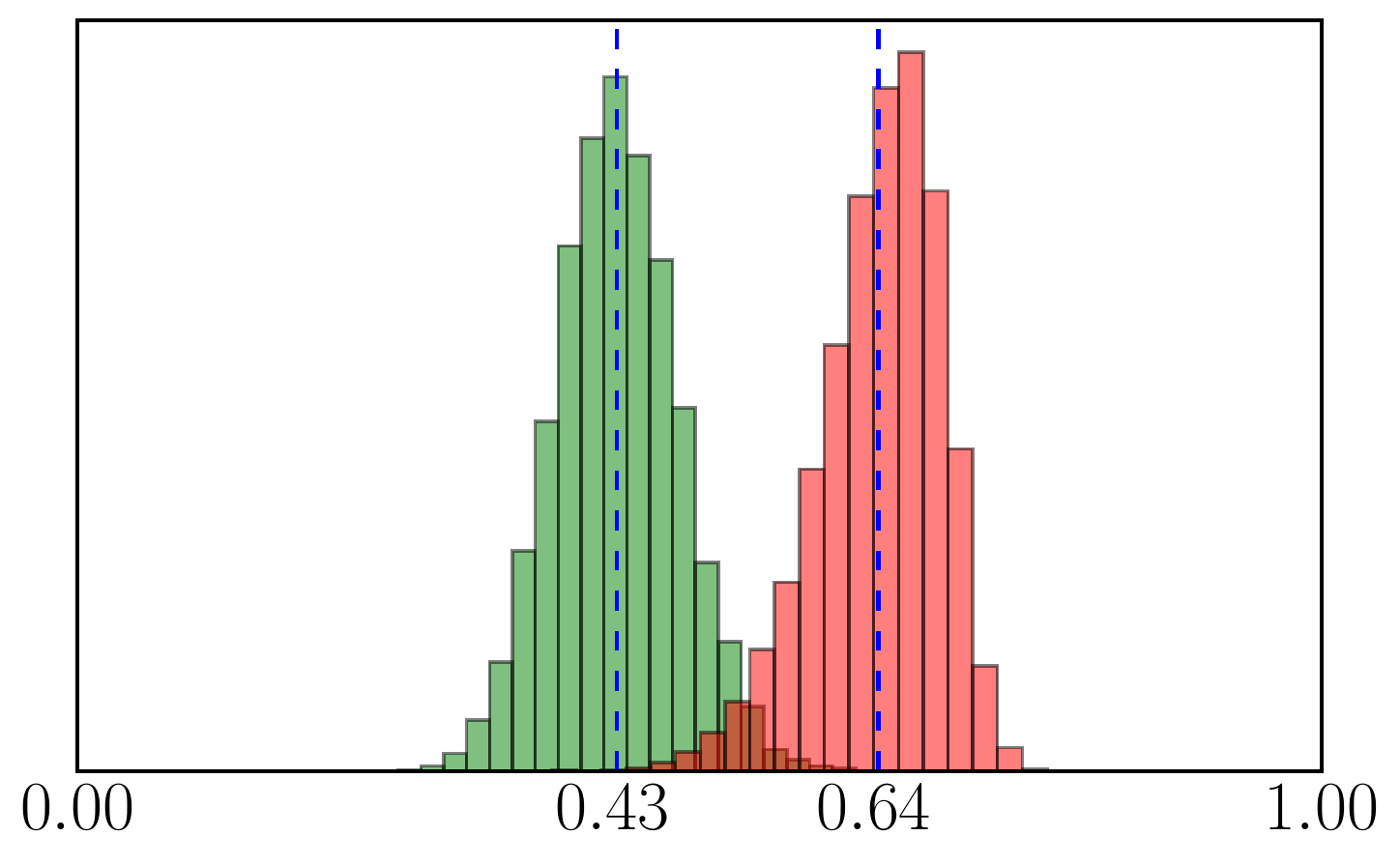}}\hfill
\subcaptionbox{\label{fig:score_distribution_b}}{\includegraphics[width=0.49\linewidth]{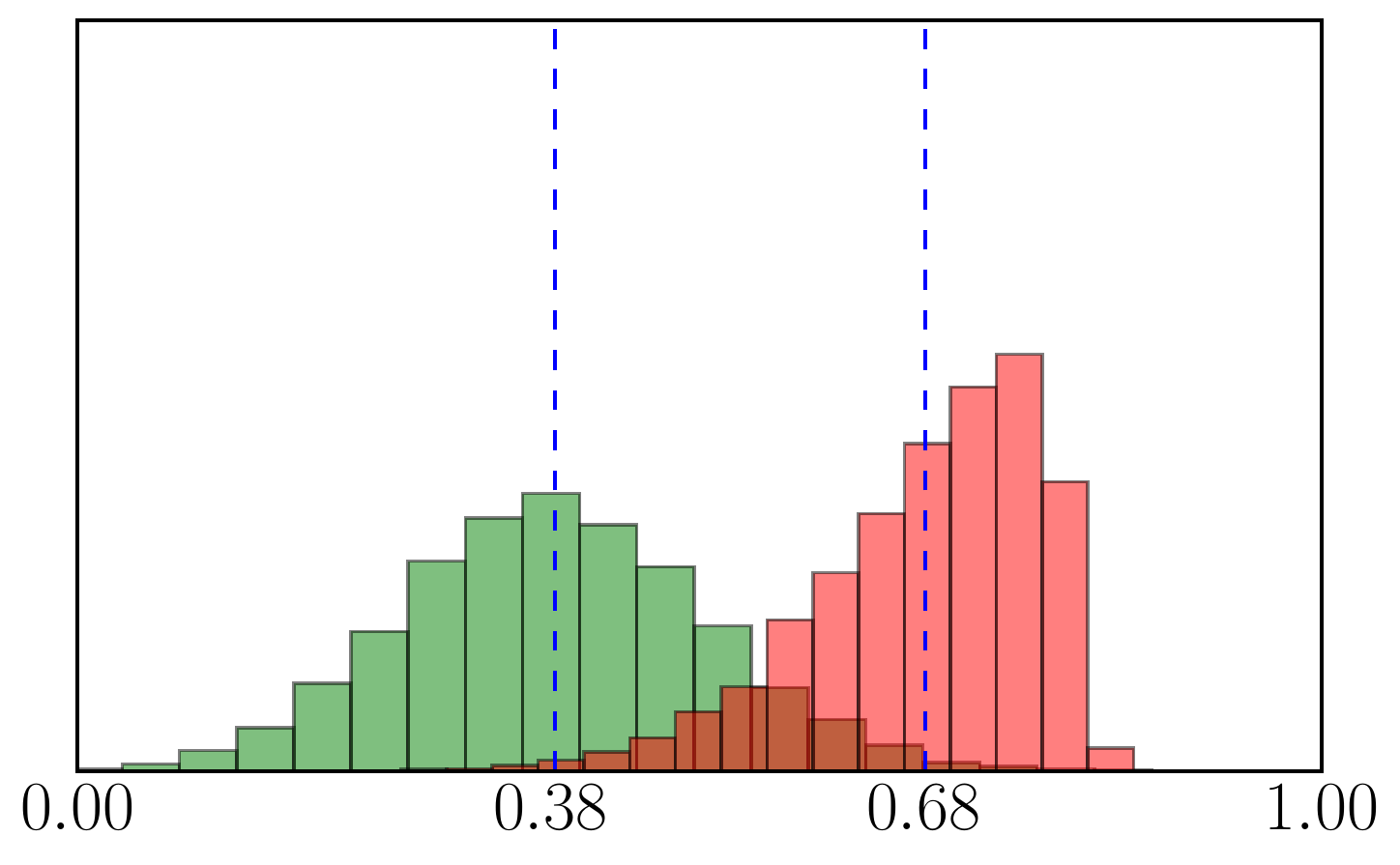}}
\caption{\subref*{fig:score_distribution_a} Distribution of positive and negative scores after training a CNN classification model. \subref*{fig:score_distribution_b} Distribution of positive and negative scores after fine-tuning the same CNN model with the standard triplet loss. Observe how even though the triplet training has been able to further separate the mean values of the two distributions, there is more overlapping between them, causing more false positives and/or false negatives}
\label{fig:score_distribution}
\end{figure}

A problem with the standard triplet loss function is that, in general, separating the mean values of the two score distributions does not ensure that the model performs well in a verification task. In \cref{fig:score_distribution} we show how a CNN model that has been fine-tuned with the standard triplet loss function is able to further separate the mean values of the two score distributions but does not produce a better accuracy. This is because there might be more overlapping between the two distributions, causing more false positives and/or false negatives. A solution to this problem is to also minimise the standard deviation of each score distribution. Our loss function is inspired by the concept of \textit{decidability}, proposed in \cite{Daugman2000Biometric} as a way of measuring the achievable accuracy of a verification system regardless of the selected threshold or operating point. A possible measure of decidability is defined as follows \cite{Daugman2000Biometric}:
\begin{equation}
d = \frac{\abs{\mu_{ap} - \mu_{an}}}{\sqrt{\frac{1}{2}\left(\sigma_{ap}^2 + \sigma_{an}^2\right)}}\label{eq:decidability}
\end{equation}
where $\sigma_{ap}^2$ and $\sigma_{an}^2$ are the variances of the distributions of positive and negative scores respectively.

\cref{eq:decidability} implies that a higher decidability $d$ is achieved by increasing the difference between the mean values of the two score distributions while decreasing both of their variances. Although it would be possible to use the inverse of \cref{eq:decidability} as our training objective, in practice, using the margin parameter $\alpha$ leads to a better separation between the two score score distributions. For this reason, we construct our loss function by adding a new term to \cref{eq:simplified_triplet_loss} that accounts for the variance in the two score distributions:
\begin{equation}
L = \left(1 - \beta\right) \left(\mu_{ap} - \mu_{an} + \alpha \right) + \beta \left(\sigma_{ap}^2 + \sigma_{an}^2\right)\label{eq:batch_triplet_loss}
\end{equation}
where $\beta$ is a parameter that balances the contribution of the two terms. In particular, at $\beta = 1$, the term that accounts for the difference between the mean values of each score distribution vanishes and only the term that accounts for the variances of the score distributions has an effect. The opposite happens when $\beta = 0$.

An advantage of adding this new term to the triplet loss function is that even if a triplet does not violate the margin condition, the loss will usually be greater than zero since the term that accounts for the variances of the score distributions is non-zero. Even though this means that adopting an online triplet sampling strategy is not strictly needed, in our experiments we noticed faster convergence when using it. Concurrent to our work, a similar loss function has been proposed in \cite{kumar2016learning} to learn local image descriptors. However, \cite{kumar2016learning} does not make use of online triplet sampling.

Note that the loss function in \cref{eq:batch_triplet_loss} cannot be expressed as the average loss for each training image since the variances need to be computed with more than one sample. Ideally, we need to train using large enough batches of images so that the variance estimation is more accurate. For this reason, we refer to this form of triplet loss as \textit{batch triplet loss}. In \cref{sub:performance_on_the_lfw_benchmark}, we show the improved accuracy when using our loss function compared to the standard triplet loss function.

% subsection batch_triplet_loss (end)

% section proposed_methods (end)

\section{Experiments} % (fold)
\label{sec:experiments}

In this section, we provide experimental results for our two contributions. In \mbox{\cref{sub:performance_on_occluded_faces}} we test different CNN models trained with occluded training images as described in \cref{sub:occlusion_maps}. We use the CASIA-WebFace database \mbox{\cite{Yi2014Learning}} to evaluate the performance on faces that present artificial occlusions and the AR face database \mbox{\cite{martinez1998ar}} to evaluate the performance on face images that present real-life occlusions. In \cref{sub:performance_on_the_lfw_benchmark} we show our experimental results on the LFW  \cite{Huang2007Labeled} benchmark using the CNN models evaluated in \mbox{\cref{sub:performance_on_occluded_faces}} and their fine-tuned versions using the standard triplet loss function and the proposed batch triplet loss function.

\subsection{Performance on Occluded Faces} % (fold)
\label{sub:performance_on_occluded_faces}

In \cref{sub:occlusion_maps} we described a training procedure for increasing the CNN model classification accuracy on occluded faces by using a probability distribution $\bm{P}$ to augment the training set with occluded training images. In this section we will start by comparing the performance of two different training schemes. The first training scheme comprises fine-tuning our baseline model A, described in \cref{sec:proposed_methods}, with occluded training images generated by sampling the occluder locations from a probability distribution $\bm{P}$. The probability distribution $\bm{P}$ is obtained by applying \mbox{\cref{eq:softmax_with_temperature}} to an occlusion map $\bm{O}$ of a particular size. Each occlusion map $\bm{O}$ was generated with model A using a subset of 1,000 images from the CASIA-WebFace validation set, as described in \cref{sub:occlusion_maps}. By contrast, the second training scheme comprises fine-tuning model A with occluded training images generated by sampling the occluder locations from a standard normal distribution. The goal of training with these two training schemes is to assess the benefits of training CNN models with images overlaid by strategically located occluders as opposed to randomly located occluders.

We train several CNN models in this manner, one for each of the occluder sizes shown in \cref{fig:map}. The temperature value $T$ in \mbox{\cref{eq:softmax_with_temperature}} was empirically set to 0.25, 0.4 and 0.6 with $\bm{O}_{20\times20}$, $\bm{O}_{20\times40}$ and $\bm{O}_{40\times40}$ respectively. We add the size of the occluder used to generate the occluded training images to the name of each fine-tuned model. Additionally, if the model was trained following the second training scheme, an R is added to the model name. For example, model A fine-tuned with occluded training images overlaid by an occluder of $20\times20$ pixels becomes $\text{A}_{20\times20}$ if the locations of the occluder are sampled from $\bm{P}$ (first training scheme) and $\text{A}_{20\times20\text{R}}$ if the locations of the occluder are sampled from a standard normal distribution (second training scheme).

\begin{table*}[tb]
\centering
\begin{tabular}{lccc}
\toprule
Model & $\boldsymbol{O}_{20\times20}$ & $\boldsymbol{O}_{20\times40}$ & $\boldsymbol{O}_{40\times40}$ \\ \midrule
A & $92.9\% \pm 10.99$ & $86.18\% \pm 18.51$ & $76.19\% \pm 27.89$ \\ \midrule
$\text{A}_{20\times20}$ & $\boldsymbol{97.69\% \pm 2.62}$ & $\boldsymbol{95.1\% \pm 5.64}$ & $\boldsymbol{88.9\% \pm 14.13}$ \\
$\text{A}_{20\times20\text{R}}$ & $97.12\% \pm 3.55$ & $93.98\% \pm 7.39$ & $86.93\% \pm 16.98$ \\ \midrule
$\text{A}_{20\times40}$ & $\boldsymbol{97.75\% \pm 2.42}$ & $\boldsymbol{95.85\% \pm 4.03}$ & $\boldsymbol{90.64\% \pm 10.88}$ \\
$\text{A}_{20\times40\text{R}}$ & $97.62\% \pm 2.9$ & $95.45\% \pm 5.12$ & $89.54\% \pm 13.29$ \\ \midrule
$\text{A}_{40\times40}$ & $\boldsymbol{98.37\% \pm 1.7}$ & $\boldsymbol{96.8\% \pm 3.16}$ & $\boldsymbol{93.47\% \pm 6.94}$ \\
$\text{A}_{40\times40\text{R}}$ & $98.31\% \pm 2.29$ & $96.52\% \pm 4.14$ & $92.61\% \pm 9.13$ \\ \bottomrule
\end{tabular}
\caption{Mean classification accuracy and standard deviation of each occlusion map $\boldsymbol{O}$ generated by different CNN models.}
\label{tab:artificial_occlusion_accuracy}
\end{table*}

To compare the accuracy of these fine-tuned CNN models we generate occlusion maps $\bm{O}$ with them (one for each of the occluder sizes). Since an occlusion map indicates the classification error incurred by a model at each spatial location, we can easily calculate the mean classification accuracy as $1-\sum_{i,j}{O_{i, j}}$. \cref{tab:artificial_occlusion_accuracy} shows the mean classification accuracy and standard deviation for each occlusion map generated with each fine-tuned model. For each model in \mbox{\cref{tab:artificial_occlusion_accuracy}}, we generated the three occlusion maps $\bm{O}_{20\times20}$, $\bm{O}_{20\times40}$ and $\bm{O}_{40\times40}$ using a subset of 1,000 images from the CASIA-WebFace validation set in such a way that all the selected images can be correctly classified by the model if no occluder is used. For example, to generate $\bm{O}_{20\times20}$, ${O}_{20\times40}$ and ${O}_{40\times40}$ with model $\text{A}_{20\times20}$, we used a subset of 1,000 images that were classified correctly by model $\text{A}_{20\times20}$. To avoid any bias in the results, we selected a different subset of 1,000 images to generate the occlusion maps used to compute the results shown in \mbox{\cref{tab:artificial_occlusion_accuracy}} and to generate the probability distribution $\bm{P}$ used when training each model. In other words, we avoided testing our models using the same images that were (indirectly) incorporated in the training stage by the use of $\bm{P}$. Observe that not only all the models fine-tuned with occluded training images achieve a higher classification accuracy than model A but their standard deviations are considerably smaller. This indicates that the performance of the fine-tuned models is much less affected by the location of the occluder, i.e. the models are able to extract discriminant features from all the  face regions more equally, regardless of the location of occluder. Moreover, the results in \cref{tab:artificial_occlusion_accuracy} show the better performance of the models trained with occluded training images overlaid by strategically located occluders compared to those trained with occluded training images overlaid by randomly located occluders.

\begin{figure}[tb]
\centering
\hspace*{\fill}
\subcaptionbox{\label{fig:ar_example_a}}{\includegraphics[page=1,width=0.25\linewidth]{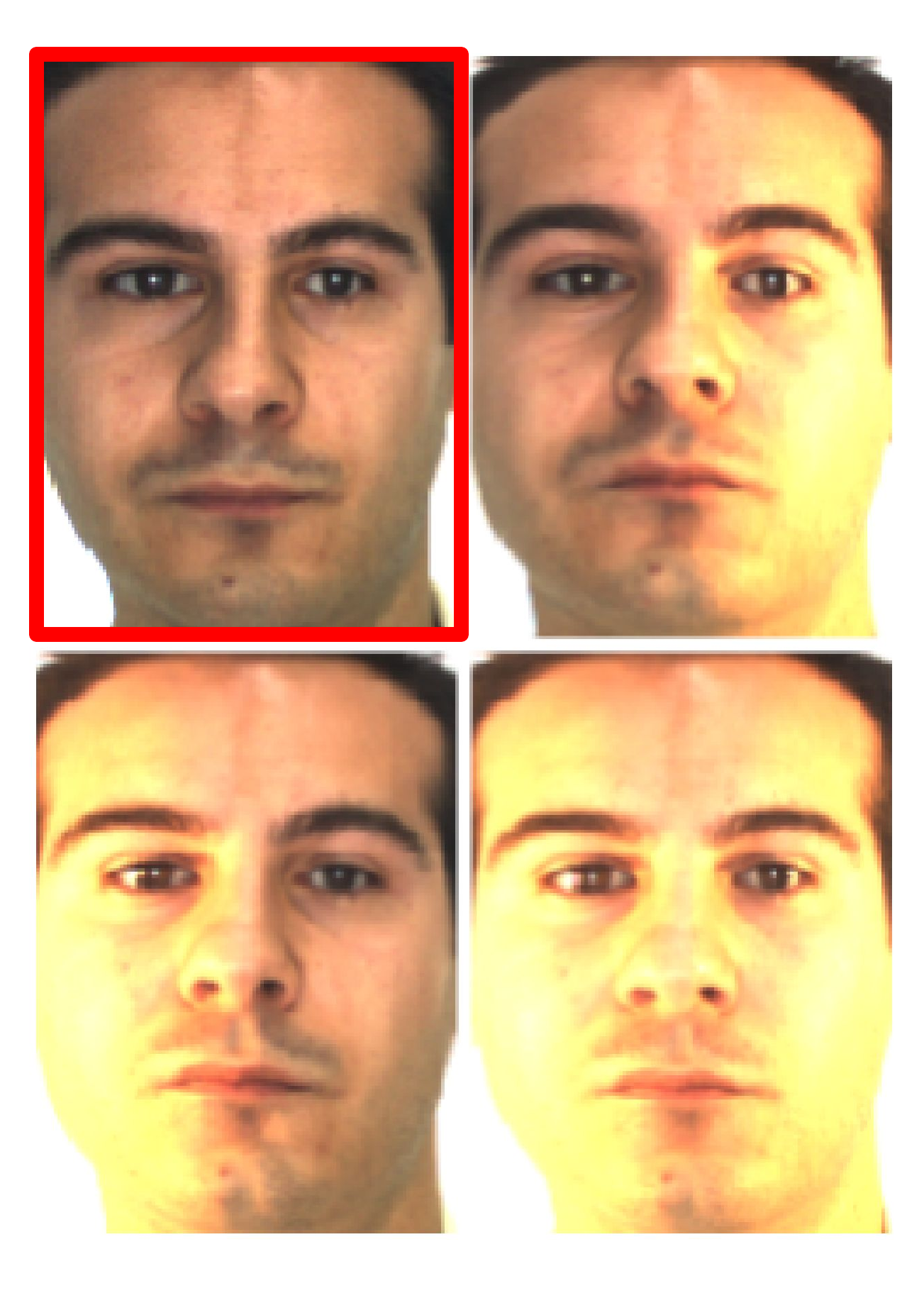}}\hfill
\subcaptionbox{\label{fig:ar_example_b}}{\includegraphics[page=2,width=0.25\linewidth]{Figures/images/fig_ar_example.pdf}}\hfill
\subcaptionbox{\label{fig:ar_example_c}}{\includegraphics[page=3,width=0.25\linewidth]{Figures/images/fig_ar_example.pdf}}
\hspace*{\fill}
\caption{Example images from the AR database. In each subfigure, the highlighted image on the top left is the reference image (target image) used to compare against the other three images (query images). \subref*{fig:ar_example_a} Non-occluded. \subref*{fig:ar_example_b} Wearing sunglasses. \subref*{fig:ar_example_c} Wearing scarf.}
\label{fig:ar_example}
\end{figure}

Since the goal is to improve the accuracy when dealing with real-life occlusions, we have further evaluated the performance of our CNN models on the AR face database \cite{martinez1998ar}. The AR face database contains 4,000 face images of 126 different subjects with different facial expressions, illumination conditions and occlusions. Out of these, we only use faces with different illumination conditions and occlusions. The different illumination conditions correspond to face images with a light on the left side, right side or both. The occluded face images consist of people wearing either sunglasses or a scarf. We carry out three different evaluations. In each evaluation, we compare non-occluded faces against (i) other non-occluded faces, (ii) faces occluded by a pair of sunglasses, and (iii) faces occluded by a scarf. \cref{fig:ar_example} shows examples of each type of image used in the three evaluations.

\begin{figure}[tb]
\centering
\subcaptionbox{\label{fig:ar_roc_a}}{\includegraphics[width=0.5\linewidth]{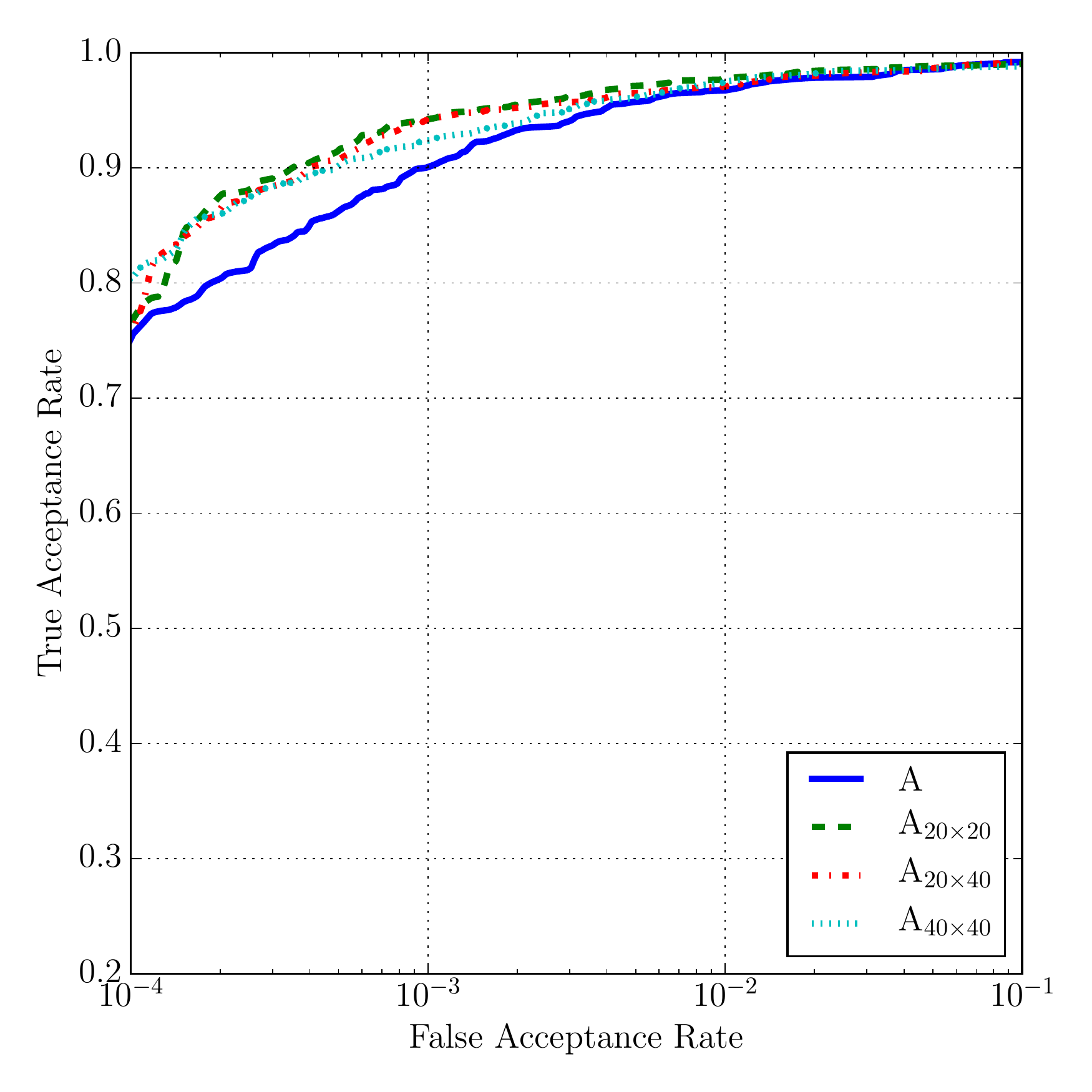}}\hfill
\subcaptionbox{\label{fig:ar_roc_b}}{\includegraphics[width=0.5\linewidth]{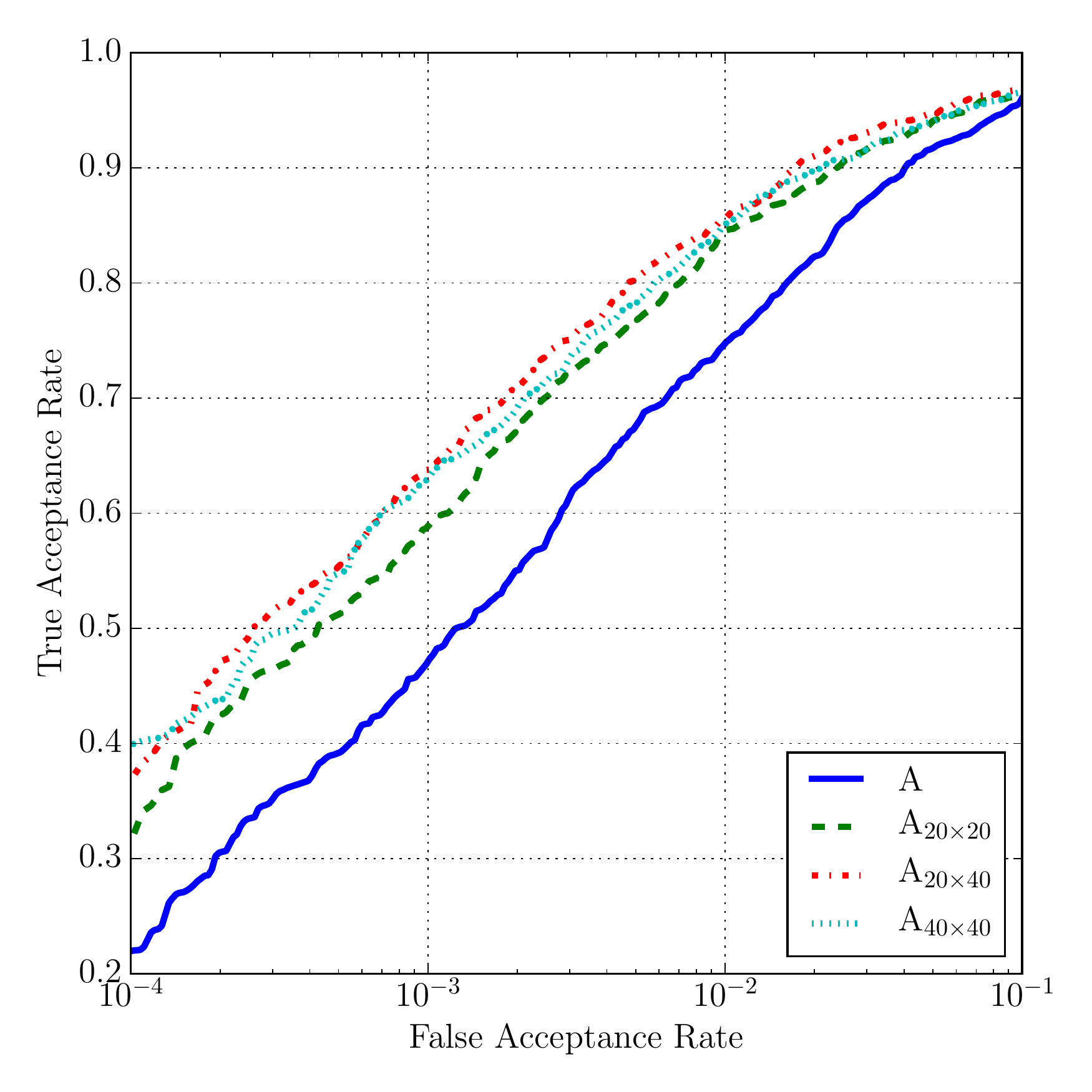}}\hfill
\subcaptionbox{\label{fig:ar_roc_c}}{\includegraphics[width=0.5\linewidth]{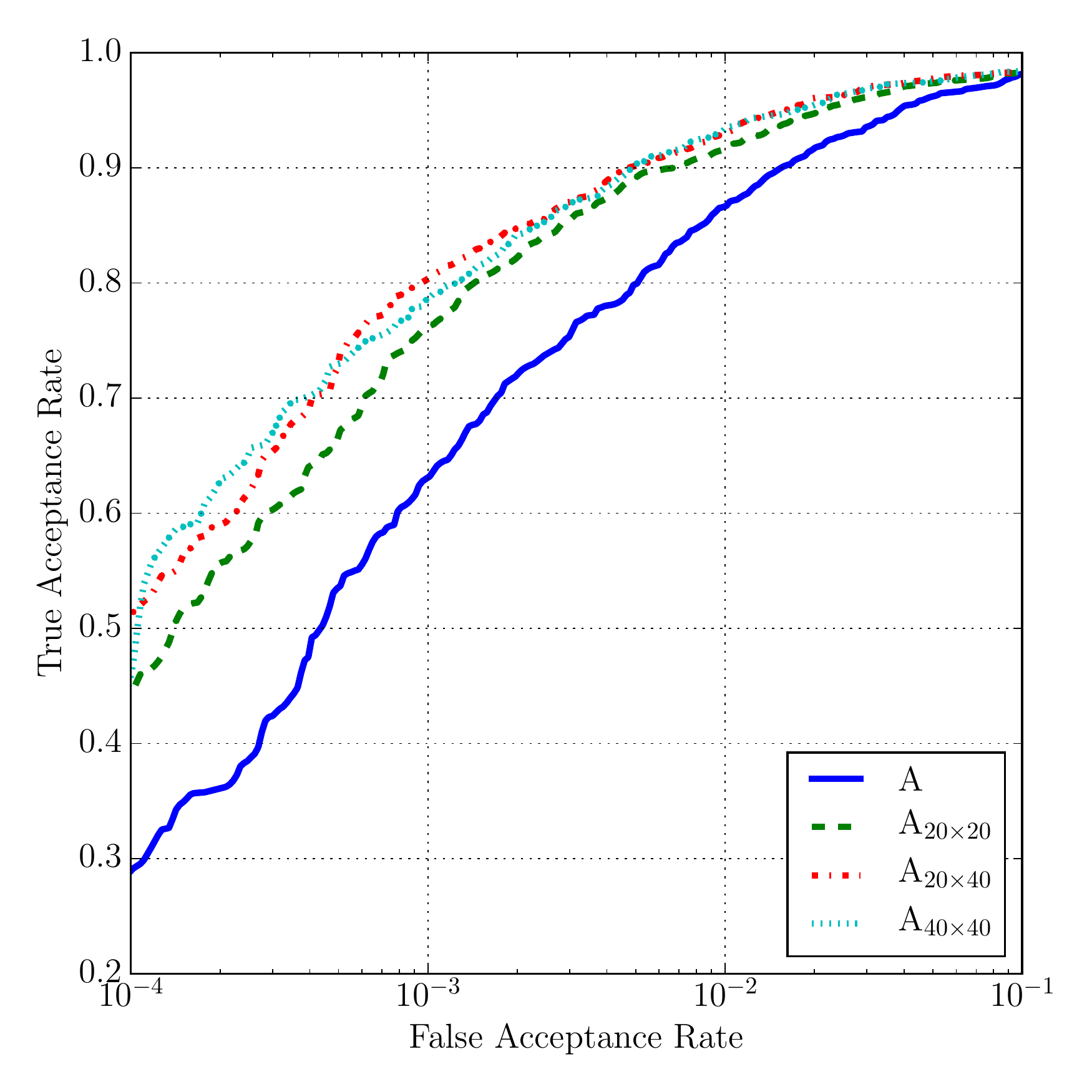}}
\caption{AR database ROC curves \subref*{fig:ar_roc_a} Non-occluded. \subref*{fig:ar_roc_b} Wearing sunglasses. \subref*{fig:ar_roc_c} Wearing scarf.}
\label{fig:ar_roc}
\end{figure}

As shown by the resulting ROC curves in \cref{fig:ar_roc_a,fig:ar_roc_b,fig:ar_roc_c}, the performance of the models trained with occluded training images consistently outperform the baseline model A, particularly at low False Acceptance Rates. Note that the performance does not seem to be greatly affected by the occluder size. The ROC curve for the evaluation of non-occluded faces (\cref{fig:ar_roc_a}) shows that model $\text{A}_{40\times40}$ performs slightly worse than models $\text{A}_{20\times20}$ and $\text{A}_{20\times40}$, perhaps because the large occluder used during training makes the model rely on fewer features. As a consequence, the model performs worse when presented with non-occluded faces in which all the face regions are visible and contain useful features. In contrast, model $\text{A}_{20\times20}$ performs worse than the other two when presented with faces occluded by a pair of sunglasses (\cref{fig:ar_roc_b}) or a scarf (\cref{fig:ar_roc_c}). This might be because the occluder used during training is too small to simulate these types of occlusions.

Observe that, even though model $\text{A}_{40\times40}$ achieved the best classification accuracy when evaluated on face images that present artificial occlusions (\mbox{\cref{tab:artificial_occlusion_accuracy}}), the results on the AR face database differ because the evaluation involves comparing pairs of occluded and non-occluded face images instead of only classifying occluded face images. For this reason, the models need to perform well not only when presented with occluded face images but also with non-occluded face images. It seems that using a medium-sized occluder like the one used to train model $\text{A}_{20\times40}$ offers the best performance when taking into account the three different evaluations, as it avoids the problems encountered with small occluders (not being able to simulate large occlusions like sunglasses and scarves) and large occluders (worse performance when presented with non-occluded faces). Note that we did not repeat these experiments with the models trained using the second training scheme described earlier, as their performance was already shown to be inferior.

% subsection performance_on_occluded_faces (end)

\subsection{Performance on the LFW benchmark} % (fold)
\label{sub:performance_on_the_lfw_benchmark}

We now adopt another approach to training by fine-tuning model A using the standard triplet loss and the batch triplet loss described in \cref{sub:batch_triplet_loss}. We do this by discarding the classification layer, normalising the features of the previous layer (bottleneck features) using the $L_2\text{-norm}$ and fine-tuning all the CNN layers with one of the two loss functions. We refer to the CNN model fine-tuned with the standard triplet loss function as model B, and the CNN model fine-tuned with the batch triplet loss function as model C. The parameter $\alpha$ is set to 0.5 when using any of these training objectives and the parameter $\beta$ is set to 0.7 when training with the batch triplet loss function. The values for both $\alpha$ and $\beta$ were obtained empirically.

Additionally, we also trained these CNN models with occluded training images. Similarly to the notation followed in \cref{sub:performance_on_occluded_faces}, we append the size of the occluder used during training to the model name. In this case, we trained these models by fine-tuning a model that has already been trained with occluded training images instead of fine-tuning model A. The locations of the occluders were sampled from the same probability distributions $\bm{P}$ that were used in \mbox{\cref{sub:performance_on_occluded_faces}}. For example, to train model $\text{B}_{20\times40}$ we fine-tuned model $\text{A}_{20\times40}$ (and not model A) with occluded training images overlaid by an occluder of $20\times40$ placed at locations sampled from the same probability distribution $\bm{P}$ that was used to train $\text{A}_{20\times40}$.

\begin{figure}[tb]
\centering
\subcaptionbox{\label{fig:lfw_example_a}}{\includegraphics[page=1,width=0.47\linewidth]{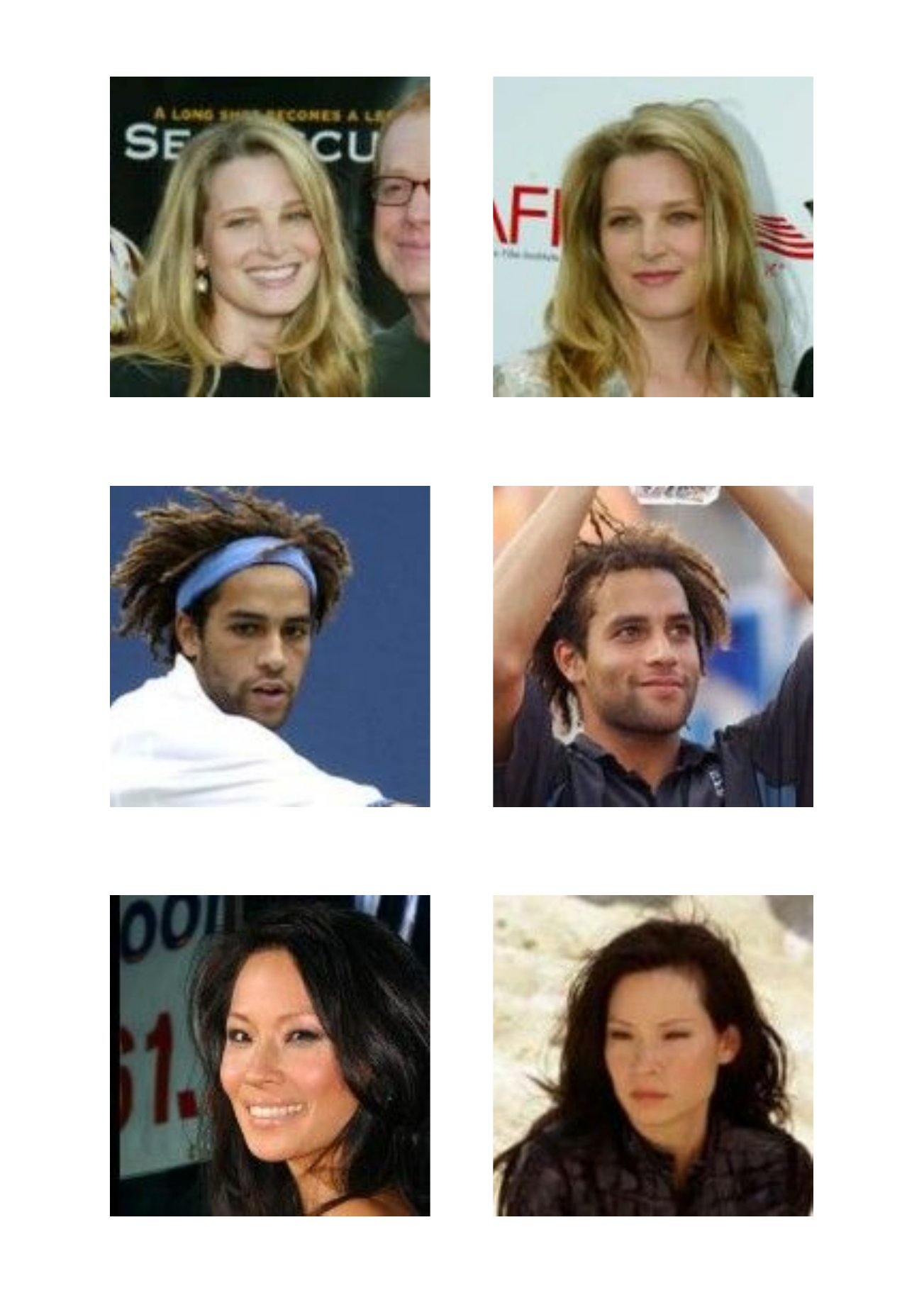}}\hfill
\subcaptionbox{\label{fig:lfw_example_b}}{\includegraphics[page=2,width=0.47\linewidth]{Figures/images/fig_lfw_example.pdf}}
\caption{Example pairs from the LFW benchmark. \subref*{fig:lfw_example_a} Matching pairs. \subref*{fig:lfw_example_b} Mismatching pairs.}
\label{fig:lfw_example}
\end{figure}

Our models are evaluated on the LFW dataset following the \textit{unrestricted, labeled outside data} protocol \cite{Huang2014Labeled} (i.e. the protocol that allows training with data that is not part of the LFW dataset). The LFW protocol divides the test set in ten splits. The classification accuracy on each test split is calculated by counting the amount of matching and mismatching pairs given a certain threshold (in our case, pairs that give a similarity score above the threshold are counted as matching pairs and pairs that give a similarity score below the threshold are counted as mismatching pairs). For each test split, we selected the threshold that gives the highest amount of correct classifications in the other nine splits. The final reported value is the mean classification accuracy and the standard deviation calculated from the ten test splits. Note that most of the face images in the LFW dataset are not occluded, therefore, we do not expect to see a performance improvement as large as that seen in our experiments with occluded faces in \mbox{\cref{sub:performance_on_occluded_faces}}. \cref{fig:lfw_example} shows examples of matching and mismatching pairs of face images from the LFW benchmark.

\begin{table}[tb] \centering \begin{tabular}{lc} \toprule Model & Accuracy \\
\midrule A & 97.33\% $\pm$ 0.71 \\ B & 97.73\% $\pm$ 0.76 \\ C &
$\boldsymbol{98.12\% \pm 0.65}$ \\ \midrule
$\text{A}_{20\times20}$ & 97.4\% $\pm$ 0.71 \\
$\text{B}_{20\times20}$ & 97.85\% $\pm$ 0.69 \\
$\text{C}_{20\times20}$ & $\boldsymbol{98.35\% \pm 0.73}$ \\ \midrule
$\text{A}_{20\times40}$ & 97.68\% $\pm$ 0.83 \\
$\text{B}_{20\times40}$ & 97.79\% $\pm$ 0.82 \\
$\text{C}_{20\times40}$ & $\boldsymbol{98.42\% \pm 0.68}$ \\ \midrule
$\text{A}_{40\times40}$ & 97.18\% $\pm$ 0.63 \\
$\text{B}_{40\times40}$ & 97.5\% $\pm$ 0.57 \\
$\text{C}_{40\times40}$ &
$\boldsymbol{98.16\% \pm 0.64}$ \\ \bottomrule \end{tabular} \caption{Mean classification accuracy and standard deviation of different CNN models evaluated following the LFW unrestricted, labeled outside data protocol.} \label{tab:lfw_accuracy} \end{table}

As shown in \cref{tab:lfw_accuracy}, all the CNN models fine-tuned with the batch triplet loss outperform the CNN models trained with the standard triplet loss, validating the usefulness of our approach. Moreover, consistent with the results shown in \mbox{\cref{fig:ar_roc}}, the CNN models trained with the $20\times40$ occluder are the best performers.

% subsection performance_on_the_lfw_benchmark (end)

% section experiments (end)

\section{Conclusions} % (fold)
\label{sec:conclusions}
We have investigated which parts of the human face have the highest impact on face recognition accuracy. The proposed occlusion maps are a good way of visualising these regions and, at the same time, provide useful information about a classification model's performance on faces that present artificial occlusions. According to our experimental results, even a state-of-the-art CNN-based face recognition model fails to maintain its high performance when these face regions are occluded (e.g. by a pair of sunglasses or a scarf). We have demonstrated how these occlusion maps can be used during the training procedure to augment the training set with face images that present artificial occlusions. These artificial occlusions are strategically positioned in locations where the performance of a CNN model trained in a conventional way is most sensitive. Training with these augmented training sets, we produce CNN models that are more robust to face occlusions. As shown in our experimental results, our proposed method has shown consistent performance improvement on face images that present artificial or real-life occlusions and on face images that do not present any occlusions.

Additionally, we have revisited the problem of learning features for a verification task using distance metric objectives. We have extended the widely used triplet loss function by adding a new term that minimises the standard deviation of the distributions of positive and negative scores. In our experiments on the LFW benchmark, the proposed batch triplet loss has consistently achieved better results than the standard triplet loss. Finally, experimental results have confirmed that the best CNN models result from a combination of our two proposed approaches, regardless of whether the face images are occluded or not.

% section conclusions (end)

\section*{Acknowledgments}
This work resulted from a collaborative research project between University of Hertfordshire and IDscan Biometrics (a GBG company) as part of a Knowledge Transfer Partnership (KTP) programme (partnership number: 009547).

% References
% \section*{References}
\bibliography{Remote,Local}

\end{document}